\documentclass[runningheads]{llncs}

\usepackage[T1]{fontenc}
\usepackage{amsmath}

\usepackage{makecell}
\usepackage{booktabs}
\usepackage[svgnames,table]{xcolor}

\renewcommand{\bf}[1]{\cellcolor{black!20}\textbf{#1}}

\usepackage{subcaption}
\usepackage{multicol}
\usepackage[numbers,sort&compress]{natbib}
\usepackage{graphicx}
\begin{document}
\title{Towards a methodology for addressing missingness in datasets, with an application to demographic health datasets}
 \titlerunning{Towards a methodology for addressing missingness in health datasets}
%
\author{Gift Khangamwa\inst{1}\orcidID{0000-0001-9318-9049} \and
   Terence L. van Zyl \inst{2}\orcidID{0000-0003-4281-630X} \and
   Clint J. van Alten\inst{1}\orcidID{0000-0002-7865-4886}}
 \authorrunning{G. Khangamwa et al.}
%
 \institute{ University of the Witwatersrand, Johannesburg, School of Computer Science and Applied Mathematics, 
 Johanneburg, South Africa \\
 \email{{ \{gift.khangamwa, clint.vanalten\}}@wits.ac.za}, \\
 url: 
 \url{http://www.wits.ac.za/csam}\\ \and
 University of Johannesburg, Institute for Intelligent Systems, Johannesburg, South Africa\\
  \email{tvanzyl@gmail.com}, \\ url: \url{www.uj.ac.za/institute-for-intelligent-systems} }
\maketitle              

\begin{abstract}
Missing data is a common concern in health datasets, and its impact on good decision-making processes is well documented. Our study's contribution is a methodology for tackling missing data problems using a combination of synthetic dataset generation, missing data imputation and deep learning methods to resolve missing data challenges. Specifically, we conducted a series of experiments with these objectives;  $a)$ generating a realistic synthetic dataset, $b)$ simulating data missingness, $c)$ recovering the missing data, and $d)$ analyzing imputation performance. Our methodology used a gaussian mixture model whose parameters were learned from a cleaned subset of a real demographic and health dataset to generate the synthetic data. We simulated various missingness degrees ranging from $10 \%$, $20 \%$, $30 \%$, and $40\%$ under the missing completely at random scheme MCAR. We used an integrated performance analysis framework involving clustering, classification and direct imputation analysis. Our results show that models trained on synthetic and imputed datasets could make predictions with an accuracy of $83 \%$ and $80 \%$  on $a) $ an unseen real dataset and $b)$ an unseen reserved synthetic test dataset, respectively. Moreover, the models that used the DAE method for imputation yielded the lowest log loss an indication of good performance, even though the accuracy measures were slightly lower. In conclusion, our work demonstrates that using our methodology,  one can reverse engineer a solution to resolve missingness on an unseen dataset with missingness. Moreover, though we used a health dataset, our methodology can be utilized in other contexts.
\end{abstract}

\section{Introduction}

Missing data is a common problem in health datasets and its impact on good decision making processes is well documented. This phenomenon of missing data plagues all scientific research endeavours generally because it is not possible to collect all data in any study, because flaws do exist in both data collection gadgets which can fail and in researchers who by their nature of being human can err. Therefore, the efforts to recover missing data, so as to have data that speaks meaningfully about any specific problem under study are a part of the scientific process.

In this paper we tackled the problem of learning from incomplete data, a problem that exists in population health datasets,  which we did in the context of machine learning. We used a demographic and health dataset which has various missing data challenges which we articulate in subsection \ref{subsec:dataset}. It is a fact that when data are missing even if one imputes all missing cases, the actual data that was missing remains unknown and hence what we create are estimates. Imputation approaches such as multiple imputation attempt to improve on the quality of the imputation value estimates by improving the process of generating the imputation value estimates \cite{Rubin1975}. 

Generally, missing data occurs in different ways; $a) $ when there is very little data due to general data unavailability or scarcity, $b) $ when there are missing values in some of the attributes in the data instances because the data was not observed, recorded or was corrupted \cite{Rubin1976,Ghahramani1994}, $c) $ or in a case when the observed dataset cannot be made available for research purposes due to privacy, security or confidentiality concerns \cite{Zheng2019}. 

However, in this study, we were interested in the first two scenarios of how missing data occurs, which can be simulated by inducing artificial missingness in a synthetic dataset.  Consequently, we embarked on a synthetic dataset generation task first, in order to have a complete dataset for use as a ground truth dataset in studying missingness and imputation methods.  We, generated a clean synthetic dataset using the Gaussian Mixture Model method based on the  parameters that were learned from the statistical distributions of the real observed datasets discussed in section \ref{sec:methods}. As a result of this, the synthetic dataset provided us with ground truth values  to use in evaluating the results of our experiments using various imputation methods. 

Moreover, part of our goal was to create a realistic dataset, so that our results on this synthetic dataset can be extended to the observed dataset. We accomplished the task of making the synthetic dataset realistic in two steps, the first being using parameters learned from the real observed dataset and the second step being ensuring that the values generated in the various features matched the original observed dataset in terms of their data type, range of values and the probability distribution. 

Therefore, in this paper we propose a methodology for tackling missing data that makes use of synthetic dataset and imputation applied to a demographic and health dataset, which we believe can be generally applicable to other datasets as well.

\subsection{Problem statement}

We tackled the problem of missing data using demographic and health survey datasets as a case study. These demographic and health survey datasets are a very important public health surveillance dataset and are invaluable in the measure and assessment of the various international development goals such as sustainable development goals. However, these datasets have various missing data problems such as; $1)$  missing values, which are indicated using codes for missing values, inconsistent values and unknown values, $2)$ varied missingness degrees in features indicated as blanks, $3)$ skewed data due to missing labels in target features such as anemia, which is an indicative feature for presence of malaria, and $4)$ finally, missing data that is missing under various missingness schemes see section \ref{sub:sec:missing:schemes}.  

Overall, these missing data problems affect the usage of such important datasets especially in building supervised predictive machine learning models that might be useful in the measurement  or forecasting of progress in attainment of sustainable development goals or indeed national health targets.


\subsection{Objectives}

The main objective of the study was to propose a methodology for tackling missing data using a demographic and health dataset as a case study. We compared different imputation methods in-order to determine the best imputation method for our dataset. Specifically, our experiments focused on the following;  $a)$ generating a realistic synthetic  dataset using the gaussian mixture model trained using parameters extracted from a cleaned subset of the observed dataset, $b)$ simulating data missingness on the synthetic data based on  prevalent missingness in the observed dataset, $c)$ recovering the missing data using state of the art imputation methods and deep learning methods, and $d)$ analysing imputation performance using accuracy of classification and imputation using direct performance metrics.  

Our hypothesis in the study was as follows; models and methods that perform well on the realistic synthetic dataset with artificially induced missingness should work equally well on our observed real dataset of interest which suffers from missingness challenges. 



\subsection{Contribution}

Our contribution in this paper is the proposition of a methodology for resolving missing data that makes use of synthetic datasets, simulated missing data degrees and missingness schemes, missing data imputation and the usage of deep neural networks as part of the missing data recovery methods. The methodology comes with a step by step approach that is described in the  sub section \ref{subsec:approach}. We believe that this methodology can be followed in order to address missingness on any novel dataset by using the building blocks that we propose herein.

\section{Background }

We begin by providing some background material related to the phenomenon of missing data, its causes and  character.

\subsection{Causes of Missing Data}

In the following subsections we highlight how missing data occurs, as earlier discussed this problem can occur under the various data missingness schemes which we get into in sub section \ref{sub:sec:missing:schemes}.

\subsubsection{Missingness -  a General Case : }

this is the most common case of missing data, where the dataset has missing components in any of its features. This type of missingness is very common in machine learning datasets such that all well known data mining methodologies such as the cross industry standard process for data mining  CRISP-DM, knowledge discovery in databases KDD and the sample explore measure model assess SEMMA, all have specific phases with stages and activities dedicated to the task of cleaning the data prior to model building. Some of the activities carried out in these phases include; making imputations, formatting data, removing redundancies and transforming the data as is relevant \cite{Marban2009,Misra2019}. 

\subsubsection{Skewed Data : } \label{skewed:data}
 
class imbalance or skewness occurs when there are more data for one particular target class than the other classes. This leads to under representation of the minority class and might affect the performance of the resultant predictive model built using such data in classifying this minority class. 

Skewness or class imbalance data problem may be addressed using sampling techniques. The sampling techniques that are mainly used to address the problem are; $a)$ over-sampling of the under-represented class by reusing the minority class data instances in the model building process to ensure that there is a balance in the number of samples with the majority class, $b)$ under-sampling the majority class so that fewer data instances from this class are used,  in order to balance numbers with the minority class \cite{Chawla2002}, $c)$ the other methods combine the two strategies such that both under-represented as well as over-represented samples are sampled.   


\subsubsection{Missing Labels : }          \label{subsec:missing:labels}

missing labels problem is another problem that is part of the data incompleteness or missingness problem.  The challenge that this causes is that though one may have a large number of data instances, the number of instances that have labels are not adequate to build effective supervised machine learning models. This problem differs from the problem discussed in the previous sub section, in that the only component missing here is the target class label from the data instance. 

\subsubsection{Data Privacy, Confidentiality and Safety Concerns : } \label{subsec:privacy}

As indicated earlier,  data may be missing even though it is available, when it is withheld due to concerns of security, privacy and confidentiality by those who own such data; in such cases synthetic data generation becomes the only avenue to study such cases see \cite{Lin2006,Anderson2014}.  

\subsection{Categories of Missing Data} \label{sub:sec:missing:schemes}

According to \citet{Rubin1988a, Rubin1976,Ghahramani1994} the phenomenon of missing data is categorized  as occurring under three different sets of circumstances, leading to categorisation of three unique missingness schemes as follows; missing completely at random MCAR, missing at random MAR and missing not at random MNAR.

Missing completely at random  Eq. \ref{eqn:mcar} occurs when there is no systematic difference between instances having missing values and those having the data; in other words the missing data is neither dependent on the observed data nor on any other unobserved data~\cite{Sterne2009,Vazifehdan2019}. This can be represented as a probability function as follows;

 \begin{equation}       \label{eqn:mcar}
 p(M|X^{obs}, X^{mis}) = p(M)
 \end{equation}
 
Missing at random Eqn. \ref{eqn:mar} occurs when there are systematic differences between instances with missing data, and those with data that can be explained by the available data, in other words the missing data is dependent on the observed data only \cite{Sterne2009,Ghahramani1994}.

 \begin{equation}                \label{eqn:mar}
 p(M|X^{obs}, X^{mis}) = p(M|X^{obs})
 \end{equation}
 
 Missing not at random Eqn. \ref{eqn:mnar} occurs when there are systematic differences between instances with missing data, and those without missing data, which cannot be explained by the available data \cite{Beaulieu-Jones2017,Sterne2009}.  This implies that, data that is missing is dependent on other missing data and not on data that has been observed \cite{Vazifehdan2019,Ghahramani1994,Rubin1975}, this third category of missingness is harder to resolve and is seldom tackled in most research work on the subject.

\begin{equation} \label{eqn:mnar}
 p(M|X^{obs}, X^{mis}) = p(M|X^{mis})
\end{equation}

\subsection{Tracking Missing Data}

In order to assess the performance of missing data recovery efforts, it is necessary to track the locations of missingness in a dataset whether it is induced or otherwise. Once missing data has been imputed this missingness indicator mechanism helps to compare imputed values against actual ground truth values if known. The same approach has been proposed and utilized by \cite{Rubin1975} and employed by numerous other researchers~\cite{Yoon2018,Yoon2020,Beaulieu-Jones2017,Ghahramani1994}. The scheme makes use of a missingness indicator matrix to track the location of missing data in a data matrix $\mathbf{X} = {\mathbf{\{x}\}  }_{1}^{n}$. As per \cite{Garcia2016}, any data matrix has two components;  a missing component $\mathbf{X} _{miss}  $  and an observed component $\mathbf{X} _{obs}  $.  A missingness indicator matrix:
\begin{equation} 
\mathbf{M_{ij} } =
    \begin{cases}
        1, & \text{  }   x_{ij}   \text{   missing}\\
        0, & \text{  } x_{ij}     \text{   observed}\\       
    \end{cases}
\end{equation}     
is used to keep a record of which data is observed and which data is not. Where for each value in the data matrix the missingness indicator matrix  keeps track of which value is available and which data value is missing using the row and column indices $i$ and $j$ for each data point $\mathbf{x}$. 
  
The missingness indicator matrix has various uses in studies that employ deep neural network methods; for instance \citet{Beaulieu-Jones2017} who used denoising autoencoders included the missingness indicator matrix in the computation of a cost function for their model. On the other hand, \cite{Yoon2018,Richardson2020} included the missingness indicator value for each feature $\textbf{m} _{j} $ as part of the training dataset in building models as shown below in Eq. \ref{eqn:missing:formula}.
  \begin{equation} \label{eqn:missing:formula}
  \dot{\mathbf{x}}  = \tilde{\mathbf{x}} \odot  \bar{\mathbf{m}} + \bar{\mathbf{x}} \odot \mathbf{m}
 \end{equation}
where $\mathbf{\mathbf{m}}$ is a missingness indicator and $\bar{\mathbf{m}}$ is the complement of the missingness indicator, $\bar{\mathbf{x}} $ is generated by a neural network, while $\tilde{\mathbf{x}}$ is the observed data point with missingness, $\odot$ is the element-wise multiplication operator these inputs generated the recovered matrix $\dot{\mathbf{x}}$   \cite{Richardson2020}.

\section{Related Work}

In this section we briefly outline some of the works that are similar to our work. Specifically, we focus on methods that are used in cases where there is missing data; we consider various missing data imputation methods and other techniques for resolving missing data cases i.e. class imbalance, missing labels and others.

We split our discussion into state of the art missing data imputations in sub-section \ref{subsec:stateoftheart:imputation}, deep learning imputation methods in \ref{subsec:deep:learning:imputation}, and finally we look at a few synthetic dataset generation methods that are used to resolve various missing data scenarios. Our approach to the discussion in this section is informed by the two main components of our study which are missing data imputation and synthetic dataset generation.

\subsection{Missing Data Imputation Methods}

There are several approaches for handling the problem of missing data. The first approach is complete case analysis whereby all data points with missing data are  ignored in the analysis. This process involves  the removal of all rows or columns that have missing data, thereafter, the clean dataset is used as a complete dataset. The other approaches implement missing data imputation which allows for filling in of missing data, the two main methods in this regard are $a)$ single value imputation approaches that impute a value once,  and $b)$ multiple imputation approaches which generate multiple imputed copies of a dataset i.e. $5$ copies. Moreover, multiple imputation allows for variability in the imputed values by varying data imputation function parameters, this is done to address the  uncertainties regarding what the unknown missing values might actually be \cite{Rubin1988a}. \citet{Misra2019} provides a good review of the state of the art methods used to address missingness in health datasets.


\subsubsection{State of the Art Methods} \label{subsec:stateoftheart:imputation}
Multiple imputation remains the major approach that all state of the art machine learning methods employ for addressing missing data. The methods include random forest which is the basis for the Missing Forest imputation algorithm which used multiple trees as estimators to generate multiple copies for missing values \cite{Misra2019}.


\subsubsection{Deep Learning Methods} \label{subsec:deep:learning:imputation}
deep neural networks have been used extensively in the health domain~\cite{mathonsi2022statistics}. One major usage of deep neural networks is in the form of deep generative models that make use of a probability distribution function $P(X)$ and generate samples $X$ from some high dimensional space  $\chi$ \cite{Wan2017}. Specifically, the generative capacity of deep neural networks makes these methods to be  very suitable for missing data imputation or data recovery. Some of the works that are relevant to our work include; the work by \citet{Richardson2020} who  devised a method called MCFlow that addressed the missing data imputation problem using a framework that combines deep neural network and Monte Carlo Markov Chains.  \citet{Beaulieu-Jones2017} made use of Denoising Autoencoders inorder to address missing data in electronic health records in a clinical trial time series dataset for ALS a progressive neurodegenerative disorder.   

Moreover, there are several works that have used generative adversarial networks, these include;
 \citet{Shang2017} who developed VIGAN a view imputation generative adversarial network for imputation of missing views in a multi view environment, where patients health data was collected from a sensor network over multiple time periods.  \citet{Lin2019} developed MisGAN, a missing data imputation method that uses an incomplete dataset to train a GAN network, which is subsequently then used for imputation purposes. A missingness indicator matrix is used to track missing data and recover imputed values from the MisGAN output dataset. \citet{Yoon2018} developed GAIN a generative adversarial imputation network that made use of missingness indicator matrix to provide a hint to the discriminator as it tried to determine which samples were imputed and which were observed in their GAN imputation framework.

Our work differs from \cite{Richardson2020,Shang2017,Lin2019,Yoon2018,Yoon2020} on the basis of the deep neural network that we used which is a denoising autoencoder. Moreover, our  methodology incorporates the use of synthetic data as a medium for training models which we intend to use in resolving missingness on an original dataset. We differ with \citet{Beaulieu-Jones2017} on both the type of dataset used and the methodological components in our approach i.e. usage of synthetic data, and the sequence of activities in our approach.




\section{Methods} \label{sec:methods}

In this section we discuss the methods that were used to accomplish the objectives of the study. Overall, in this study, the use of a realistic synthetic dataset and the simulation of missingness were key to our methodology. This meant that the generation of the synthetic dataset needed to be based on parameters generated from the real dataset with missingness, likewise the artificially induced missingness needed to be similar to the actual missingness of the observed dataset in both degree and scheme of missingness. Moreover, in our approach in sub-section \ref{subsec:approach}, we ensured that artificial targets were generated using a DNN model that was exposed to the original targets. This was done so that these generated artificial targets in the synthetic dataset should be similar and connected to the original targets.

Our discussion is structured as follows; we discuss our datasets in subsection \ref{subsec:dataset}, thereafter we discuss the approach that was taken in a step by step manner in subsection  \ref{subsec:approach}, and finally we discuss the metrics of interest that we use to present our work in subsection \ref{subsec:metrics}.

\subsubsection{Dataset Description} \label{subsec:dataset}

We used the latest DHS survey datasets for the Southern African countries of Malawi, Namibia, Zambia and Zimbabwe. We used these datasets in-order to learn the parameters for generating our synthetic datasets. These countries were selected because they are within the  Southern Africa region of Sub-Saharan Africa in the malaria endemic region and have data on malaria, anemia, diabetes and hypertension, which are among the biggest health challenges for the region and the continent~\cite{manaka2022using}. 

The observed datasets have several data incompleteness problems such as: missing values, which are indicated using codes for missing values, inconsistent values and unknown values; blanks; skewness and missing labels for select target features.  Therefore, we selected a file from these observed datasets and cleaned it to have instances that have features in a select set of features only. We thereafter used this cleaned dataset to generate a synthetic dataset as described in the approach in section \ref{subsec:approach}. Overall, the dataset had a  missingness rate of about $23.8 \%$

\subsection{Approach} \label{subsec:approach}

In this subsection we outline the steps followed in our approach to the study and the specific methods used in the following sequential steps.

\begin{figure}[htb!]
\centering
\includegraphics[width=0.9\linewidth]{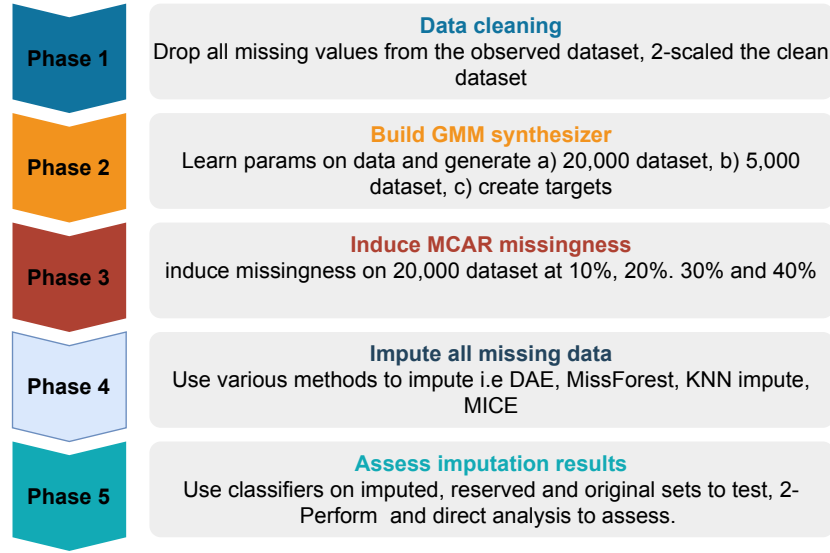}
\caption{A phased summary of the steps in the approach taken.
} 
\label{fig:approach}
\end{figure}

\begin{enumerate}
\item Our experiment begun by loading a copy of the  demographic and health dataset a survey dataset that we selected as our study dataset. 
Thereafter, we dropped all missing values to create a clean sub-dataset having most of the relevant anemia features. The size of the clean dataset was $2,058$ instances and $56$ features.  

\item Afterwards,  we scaled the clean sub-dataset using a minimax scaler to avoid some features from having a dominating effect on the synthetic dataset generation models. Our objective was to ensure that our data be generated by  different components of a Gaussian Mixture Model data generator.

\item Next, we used a Gaussian Mixture Model (GMM) to learn hyper parameters for our synthetic dataset generator models. First of all, we performed a parameter search to identify the ideal  number of components and an appropriate  covariance matrix shape. In this regard, during the search we compared the models using Akaike Information Criterion (AIC) and Bayesian Information Criterion (BIC) inorder to pick the best models. Both AIC and BIC are good tools for model selection, they are defined as given in equations Eq. \ref{eqn:aic} and Eq. \ref{eqn:bic}. Subsequently, we used the parameters of the best models from the previous steps in the final GMM synthetic dataset generator model that we used to generate a dataset having $20,000$ data points. We also created a testing dataset with $5,000$ data point which we reserved to test our models later in step $7$.

\item We thereafter created targets for the synthetic dataset using partially trained Deep Neural Network (DNN) models. The DNN models were trained using the scaled clean datasets from step $2$. We chose a DNN design with two deep layers having $20$ nodes each and one dropout layer at $20 \%$ dropout rate. Our strategy to avoid over-fitting included using both early stopping and the $20 \%$ dropout layer in the training of the target generator model. 
Moreover we made sure to use all the data points in the training of the generator. 
We also used model check-pointing to identify and save the best DNN model from the training step. Thereafter, the best DNN model was used to predict labels for the synthetic dataset that was generated using GMM and had $20,000$ data points see  step $3$. Targets were also created for our reserved testing dataset with $5,000$ data points using the same approach.  

\item The next step was to induce missing completely at random (MCAR) missingness at different degrees in the main synthetic dataset i.e. the one having $20,000$ data points, we induced missingness degrees ranging from  $10 \%$, $20 \%$, $30 \%$,  and, $40 \%$ missing values. Each combination of missingness scheme and missingness degree yielded a dataset with missing values. For each missingness dataset we kept track of where exactly in the dataset was the missingness  induced using a missingness masking matrix i.e. missingness indicator matrix. 

\item In the next step, we proceeded to impute missing values using four methods namely; denoising autoencoder (DAE), random forest based MissForest, k nearest neighbour (KNN) impute and MICE (iterative imputer under Multiple Imputation). The imputation process was carried out five times for each of the datasets having missingness from the previous step $6$ above.

\item In the final step, we assessed the outputs of the imputation step using classification and direct analysis of imputation. 

In the classification analysis we used a DNN custom classifier model having a similar structure to the target generator model. 
 Moreover, we repeated our experiments atleast $10$ times to get mean performance metrics. 
 Specifically, to train our models we used imputed datasets that were split into a training and validation set. On the other hand, in order to test our models we used datasets that were not used in the training steps at all, we used these datasets  $a)$  the clean original dataset from step $1$, $b)$ the full synthetic dataset without missingness from step $3$,  and $c)$ the reserved  synthetic test dataset having $5000$ data instances. 

In the direct assessment of imputation we used the missingness indicator (masking matrix) to locate all missing data locations in the dataset and compared imputed versus actual values. We used various accuracy and error measures such as accuracy, root mean square error, mean absolute percent error and other metrics in our assessment of the imputation results. Finally, in Figure \ref{fig:approach} below we provide a graphical summary of the main steps in our approach.

\end{enumerate}

\subsection{Metrics of Interest} \label{subsec:metrics}
Our first two metrics of interest were accuracy and the binary cross entropy also referred to as log loss. Accuracy is computed based on confusion matrix output that yields true positives, true negatives, false positives and false negatives. Accuracy as a metric has its weaknesses especially if a dataset is skewed, however, it is a well accepted and intuitive metric. Log loss computation hinges on true value $y_{i}$ and predicted value $p(y_{i})$ and the number of samples $N$. Using these two metrics we can assess any classification model. 

The next set of metrics are mean absolute percent error, and root mean squared error which were used to directly assess the outcome of imputation. In these metrics we consider features in the synthetic dataset as the true values ${y}_{i}$, while the imputed values are the predicted values $\hat{y}$.

The next metrics are Silhouette score Eq. \ref{eqn:silhouette:score} and Rand score, which offer  the opportunity to review imputation using KMeans clustering. The silhouette score is based on two distance measures between each sample in relation to other samples in its own cluster and the next cluster and is computed as follows;

\begin{equation} \label{eqn:silhouette:score}
    \text{Silhouette score} = \frac{b - a}{ max(a, b)}
\end{equation}

 where $a$ is mean distance to other samples in the same cluster, and $b$ is the mean distance to samples in the next cluster. In this experiment we used Euclidean distance to measure these distances.
 
 On the other hand the Rand score compares how effectively the clustering allocates the samples based on their class membership, where ideally members of one class must also be clustered together. Computation of this score is similar to that of accuracy. Where in this case $a$ and $b$ are predicted correctly as belonging to their groups, similar to the true positives and true negatives in a binary classification confusion matrix, while $c$ and $d$ are the incorrect predictions. 
 
 \begin{equation} \label{eqn:rand:score}
     Rand = \frac{a + b}{ a + b + c + d} = \frac{TP + TN}{TP + TN + FP + FN}
 \end{equation}
 
 The method requires true and predicted clusters, where the true clusters are given by the class memberships that exist in the dataset and the predicted are what the cluster algorithm generates.

Finally, metrics Akaike Information Criterion Eq. \ref{eqn:aic}, and Bayesian Information Criterion Eq. \ref{eqn:bic} were used to determine the best model based on number components for the GMM in the synthetic dataset generation experiment.

\begin{equation} \label{eqn:aic}
    AIC = \frac{-2}{N}* LL + 2 * \frac{k}{N}
\end{equation}

\begin{equation} \label{eqn:bic}
    BIC = -2 * LL + log(N) * k
\end{equation}

where $N$ is the number of training instances, $k$ is the number of parameters of the model i.e number of components in our case, and $LL$ is the log likelihood of the model. Moreover, BIC as can be seen above penalizes model complexity.

\section{Results and Discussion} \label{sec:results}

\subsection{Synthetic Data}
\begin{figure}
\begin{subfigure}[b]{0.49\textwidth}
\centering
\includegraphics[width=\textwidth]{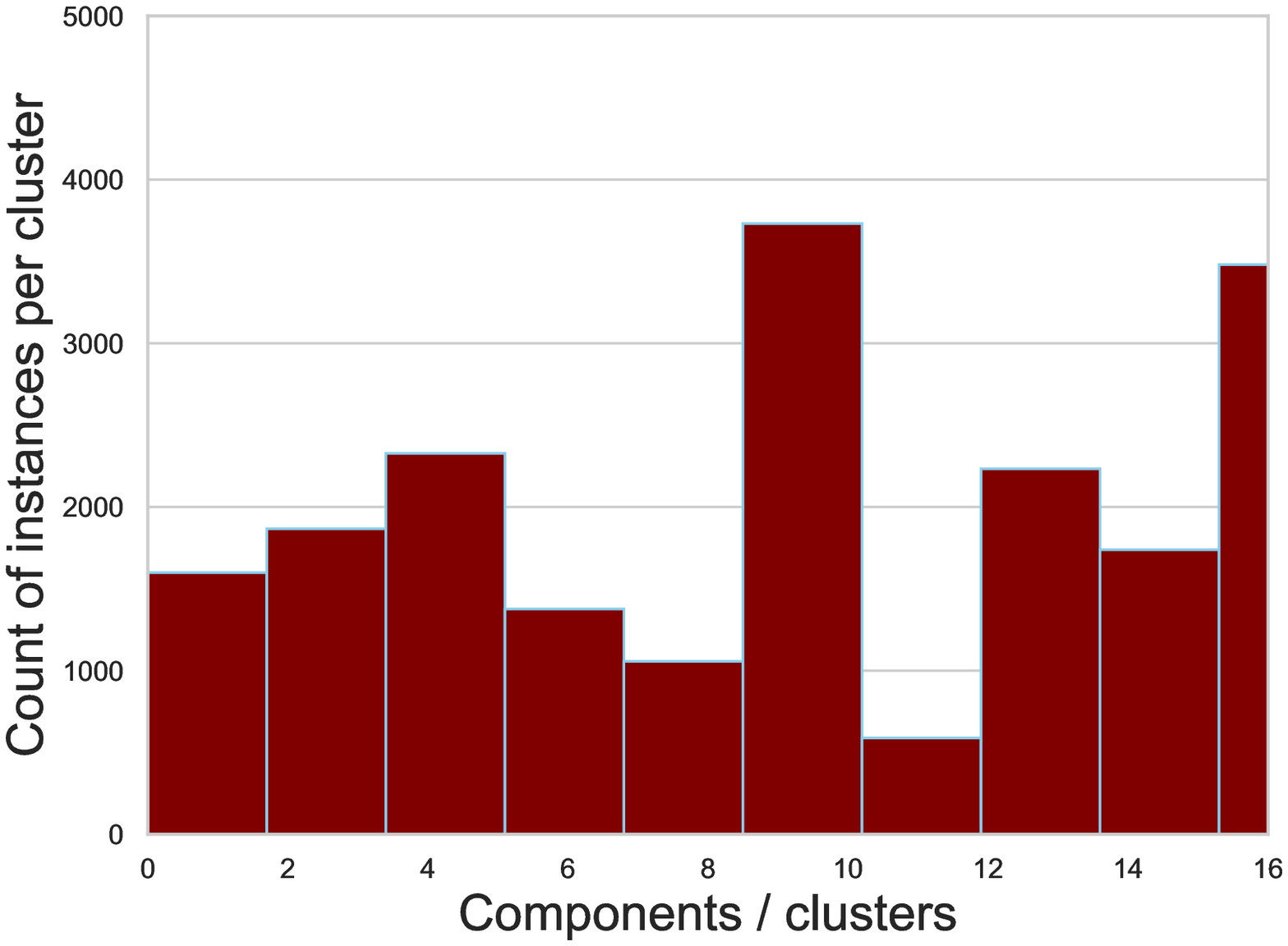}
\caption{Gaussian mixture model data generation showing data points generated per component 
} 
\label{fig:gmm:components}
\end{subfigure}
\hfill
\begin{subfigure}[b]{0.49\textwidth}
\centering
\includegraphics[width=\textwidth]{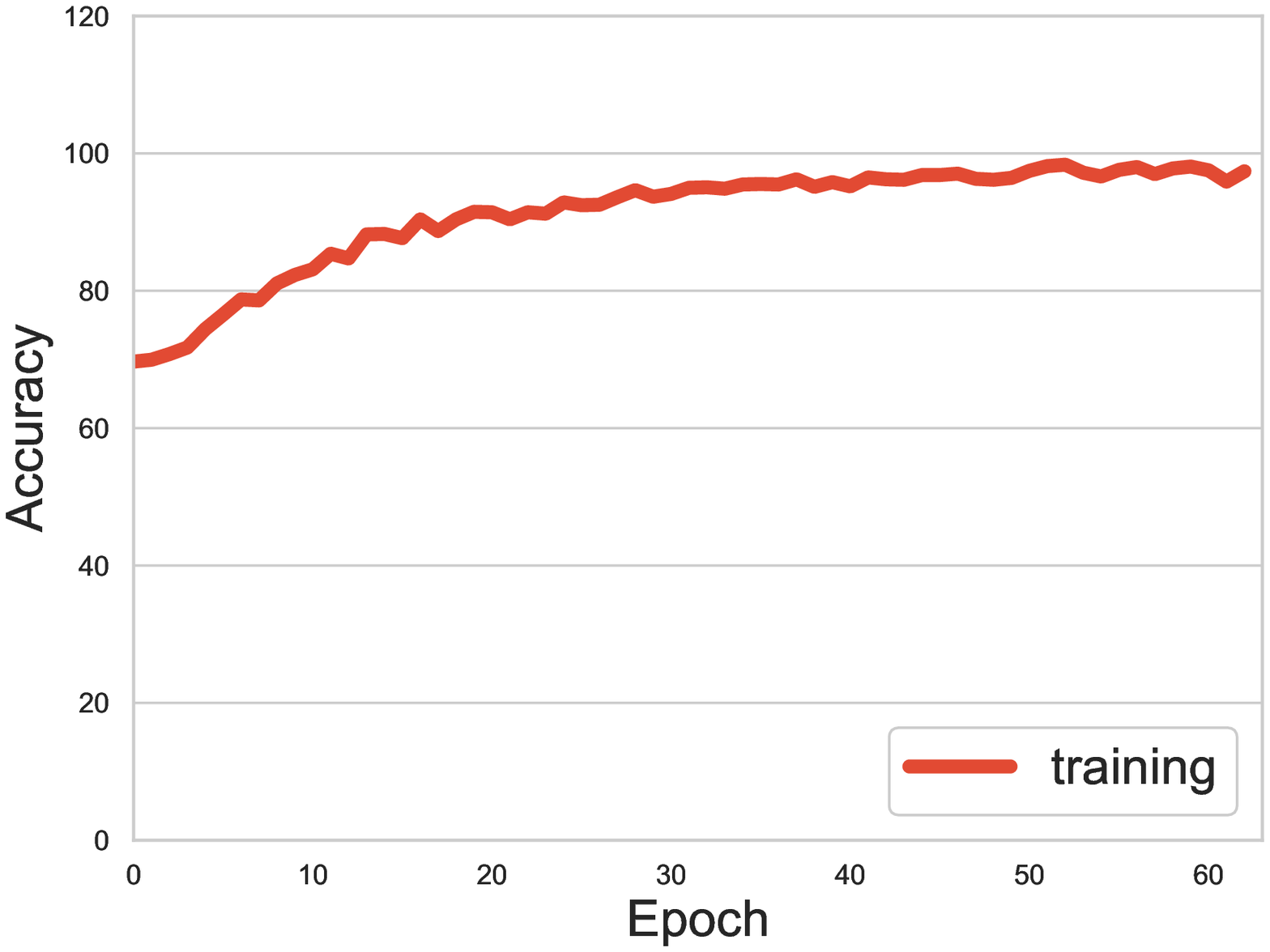}
\caption{Target generator training accuracy after partial training. Training done using the clean original sub dataset} 
\label{fig:target:gen:accuracy}
\end{subfigure}
\end{figure}

The Figure \ref{fig:gmm:components} shows the results of the synthetic dataset generation, the Figure visualizes the distribution of the generated synthetic samples by component. Moreover, once the synthetic data generation was complete, the next step was the target generation which was done using a DNN.  Figure \ref{fig:target:gen:accuracy} shows the accuracy obtained in the partial training of target generator as per the methodology described in sub-section \ref{subsec:approach}.

\begin{figure}[htb!]
     \centering
     \begin{subfigure}[b]{0.5\textwidth}
         \centering
         \includegraphics[width=\textwidth]{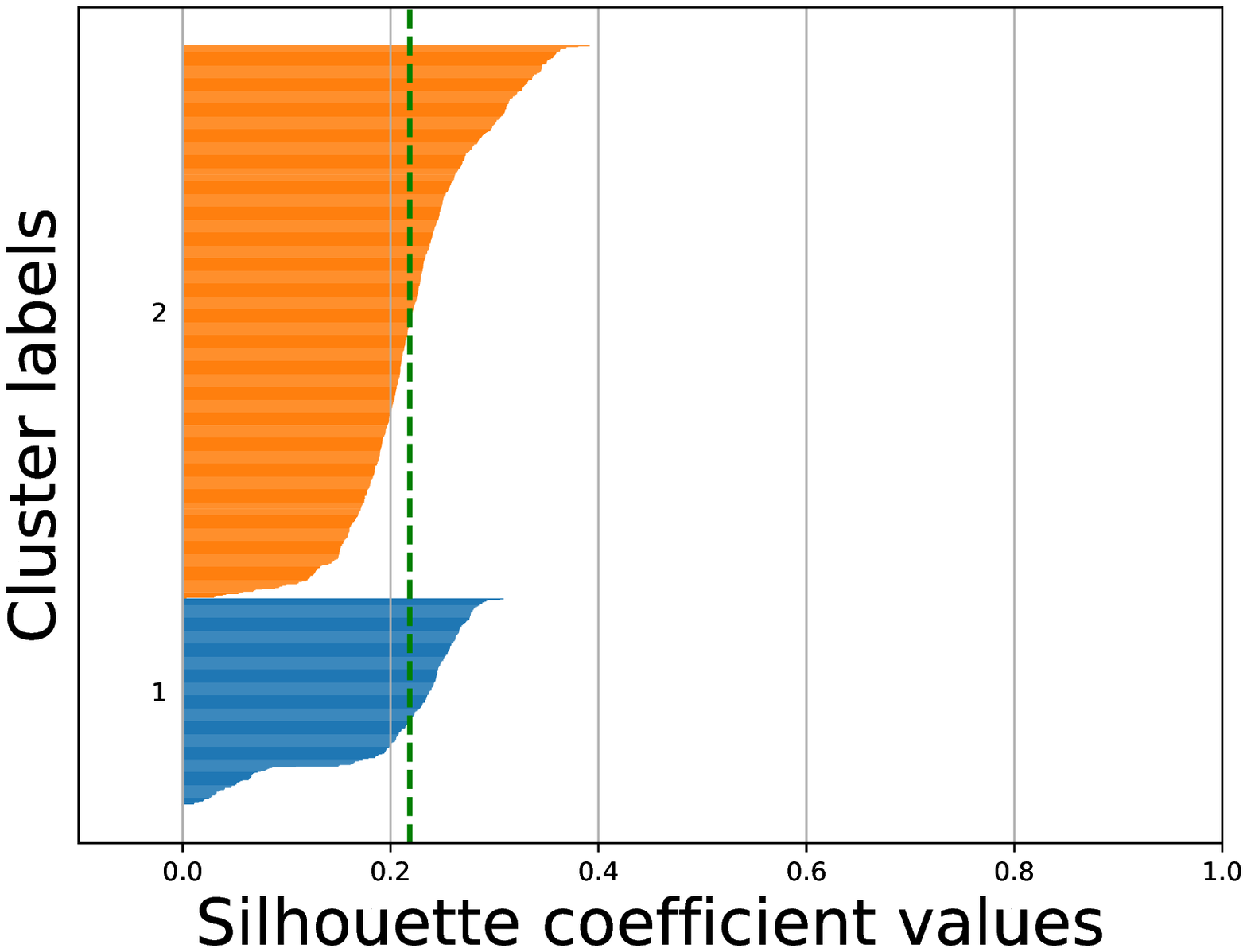}
         \caption{KMeans silhouette score analysis for the clean original dataset}
         \label{fig:original:silhoute:kmeans:2}
     \end{subfigure}
     \begin{subfigure}[b]{0.5\textwidth}
         \centering
         \includegraphics[width=\textwidth]{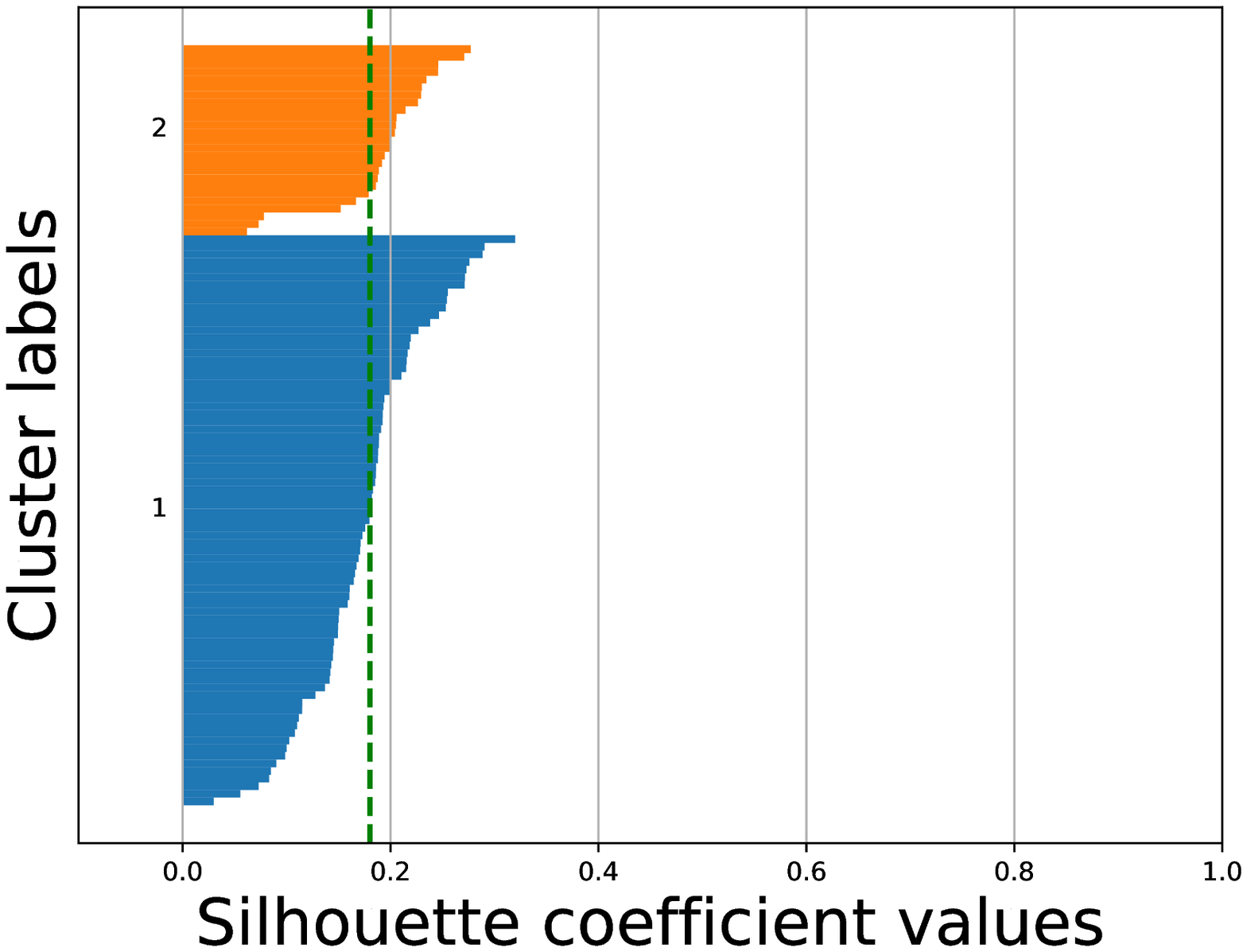}
         \caption{KMeans silhouette score analysis for the synthetic dataset}
         \label{fig:synthetic:silhoute:kmeasn:2}
     \end{subfigure}
     
\end{figure}

In the images below we present a visual comparison of the synthetic dataset generated data and the clean original sub-dataset using a K-Means 2 cluster analysis. Specifically in this analysis we plot the Silhouette score for the original clean dataset in Figure \ref{fig:original:silhoute:kmeans:2} and the synthetic dataset in Figure \ref{fig:synthetic:silhoute:kmeasn:2}  in order to have a visual analysis of the goodness of the synthetic dataset in comparison to the original. We note that in both plots there is a skew or imbalance favouring one class.

\subsection{Classification}

First of all, in Table \ref{table:classification:accuracy}, we present the classification accuracy of the imputed datasets, we also include the imputation accuracy of the full synthetic dataset as our baseline scores. The table presents the training scores given in columns $3$ and $4$ labelled training and validation respectively. On the other hand testing of the models was done on novel datasets i.e. novel to the training process of the model. Specifically, we tested the models using the full synthetic dataset, the original clean sub-dataset, and the original clean sub-dataset that was balanced using the SMOTE Edited Nearest Neighbour in columns $5$, $6$, $7$ and $8$ respectively.

\begin{table}[htb!]
\centering
\caption{Classification accuracy for models built using imputed datasets and tested on the three testing datasets a) full synthetic, b) original clean sub-dataset and balanced original using the SMOTE Edited Nearest Neighbour}
\label{table:classification:accuracy}
\resizebox{\columnwidth}{!}{%
\begin{tabular}{lrrrrrrr}
\toprule 
 \textbf{Method} &  
 \textbf{Missing} &  
 \textbf{Training} &   
 \textbf{Validation} &  
 \textbf{Synthetic} &  
 \textbf{Testing} &  
 \textbf{Original} &  
 \textbf{\makecell[cr]{Edited NN}} \\

\bottomrule
\toprule
 DAE &       10 &     87.13 &       73.75 &           75.57 &              71.06 &                    75.51 &               53.11 \\   
DAE &       20 &     87.14 &       73.76 &           74.91 &              71.08 &                    75.83 &               41.25 \\
 DAE &       30 &     82.46 &       65.82 &           65.10 &              63.24 &                    72.74 &               62.26 \\
 DAE &       40 &     87.08 &       73.92 &           74.93 &              70.68 &                    77.55 &               55.25 \\

\\
MICE &      10 &     93.88 &       83.47 &           83.88 &              79.11 &                    80.08 &               52.72 \\    
MICE &       20 &     93.71 &       84.02 &           82.53 &              78.51 &                    80.91 &               63.42 \\ 
II &       30 &     90.42 &       86.50 &           72.10 &              70.33 &                    83.89 &               67.12 \\    
         
MICE &       40 &     88.00 &       80.78 &           74.22 &              68.95 &                    78.94 &               59.14 \\   

  \\
KNN &       10 &     91.05 &       77.25 &           80.11 &              75.67 &                    83.83 &               51.17 \\

KNN &       20 &     89.12 &       75.14 &           78.34 &              73.51 &                    79.31 &               48.93 \\

KNN &       30 &     88.09 &       73.94 &           76.60 &              73.25 &                    78.18 &               55.45 \\
KNN &       40 &     86.89 &       72.12 &           75.14 &              73.09 &                    78.64 &               64.98 \\

  \\
 Miss F &       10 &     91.92 &       80.01 &           80.21 &              74.48 &                    81.01 &               49.81 \\

Miss F &       20 &     91.13 &       79.69 &           79.86 &              75.15 &                    81.64 &               46.79 \\

Miss F &       30 &     89.69 &       79.39 &           78.51 &              75.13 &                    80.37 &               56.42 \\
Miss F &       40 &     88.38 &       78.73 &           77.42 &              75.07 &                    81.47 &               54.96 \\

      \\
  Synthetic &   None &            96.28 &       91.12 &           NA &              88.57 &                    82.66 &               45.91 \\
  
\bottomrule

\end{tabular}}
\end{table}

Secondly, Table \ref{table:classification:loss} presents the training loss which was computed using the log loss as defined in sub-section \ref{subsec:metrics}. The table columns are organised as in the accuracy Table \ref{table:classification:accuracy} as described above, however, in this table we focus on the classification loss for all the imputation methods and the baseline synthetic dataset.

\begin{table}[htb!]
\centering
\caption{Classification loss for models built using imputed datasets and tested on the three testing datasets a) full synthetic, b) original clean sub-dataset and balanced original using the SMOTE Edited Nearest Neighbour
}
\label{table:classification:loss}
\resizebox{\columnwidth}{!}{%
\begin{tabular}{lrrrrrrr}
\toprule
 \textbf{Method} &  
 \textbf{Missing} &  
 \textbf{Training} &  
 \textbf{Validation} &  
 \textbf{Synthetic} &  
 \textbf{Testing} &  
 \textbf{Original} &  
 \textbf{Edited NN} \\
    \bottomrule
    \toprule
DAE &       10 &     0.088 &       0.318 &           0.142 &              0.644 &                     0.654 &               6.418 \\

      DAE &       20 &     0.082 &       0.323 &           0.159 &              0.713 &                     0.676 &              10.073 \\

      DAE &       30 &     0.240 &       0.972 &           1.844 &              2.697 &                     1.154 &              11.920 \\
      DAE &       40 &     0.073 &       0.339 &           0.197 &              0.859 &                     1.154 &              11.333 \\

  \\
MICE &       10 &     0.146 &       0.661 &           1.143 &              1.467 &                     0.774 &               8.813 \\
      MICE &       20 &     0.151 &       0.645 &           1.110 &              1.467 &                     0.751 &               7.239 \\
     MICE &       30 &     0.224 &       0.374 &           3.026 &              3.227 &                     1.617 &               7.599 \\
     MICE &       40 &     0.250 &       0.601 &           8.483 &              8.296 &                     0.804 &              13.043 \\
  \\
KNN &       10 &     0.193 &       0.940 &           1.642 &              2.243 &                     1.570 &               7.686 \\

     KNN &       20 &     0.236 &       0.958 &           1.577 &              2.220 &                     0.783 &              11.811 \\

      KNN &       30 &     0.254 &       1.018 &           2.080 &              2.776 &                     0.904 &              13.562 \\
      KNN &       40 &     0.283 &       1.027 &           2.282 &              2.917 &                     0.866 &              10.888 \\

  \\
 Miss F &       10 &     0.166 &       0.792 &           1.395 &              2.192 &                     0.754 &               9.514 \\
  
      Miss F &       20 &     0.185 &       0.818 &           1.644 &              2.549 &                     0.793 &               9.197 \\
  
     Miss F &       30 &     0.221 &       0.714 &           1.531 &              2.179 &                     0.820 &               9.365 \\
     Miss F &       40 &     0.248 &       0.743 &           1.973 &              2.787 &                     0.833 &              13.938 \\
           \\
  Synthetic &   None &             0.097 &       0.350 &      NA &    0.535 &    10.050 &      8.155 \\
    \bottomrule
    \end{tabular}}
\end{table}

Overall, our classification results presented in Table \ref{table:classification:accuracy} and Table \ref{table:classification:loss} show that the models from the imputed datasets yielded classifiers that were able to give a higher performance on the yet unseen test clean sub-dataset as well as the original cleaned observed sub-dataset. Moreover, the DAE imputed methods yielded the lowest log loss in training and also on the other test dataset, an indication of good performance. This shows that the imputation from this process successfully managed to learn from the synthetic dataset a set of parameters i.e. weights which though they represent the synthetic dataset with missingness, some of that knowledge can be transferable to the original dataset with missingness.  The other two state of the art missing data imputation methods the KNN imputer and the Miss Forest imputer came second and equally performed well especially on the test dataset and the full synthetic dataset. It should be noted that though the  performance on the original dataset can be improved accross all these methods, we have  demonstrated the potential of our approach of using synthetic dataset as an indirect route to resolve missingness in a real observed dataset.

\subsection{Direct Analysis of Imputation}

In this subsection we present results of the direct analysis of the imputation. We begin by providing results of KMeans clustering of the imputed datasets across all missingness degrees in Table \ref{table:rand:silhoutte:scores}. The table contains Rand and Silhouette scores for cluster scenarios i.e. $2$, $3$ and $4$. In both metrics a score of $1$ is the best while $0$ is the worst. 

Our results in Table \ref{table:rand:silhoutte:scores} show that the DAE imputed datasets yielded better clusters on the Silhouette score, while on the Rand score performance varied across the imputed methods. This means that the DAE yielded better clusters and yielded a better approximation of the groups in the dataset for all cluster scenarios. While the same DAE did not emerge a winner on the RAND score that is dependent on actual class label information. However, the scores between the methods on these metrics were not significantly different especially for the $2$ cluster scenario, showing that all the methods managed to recover the binary class nature of the dataset.

\begin{table}[hbt]
\centering
\caption{Rand and Silhoutte scores for various cluster sizes on imputed datasets with 30 pct induced missingness, compared with the full synthetic and original clean sub dataset.}
\label{table:rand:silhoutte:scores}
\resizebox{\columnwidth}{!}{%
\begin{tabular}{llrrrrrrrrrrrrrr}
    \toprule
 &&&&\multicolumn{5}{c}{\textbf{Rand scores}} &&&
 \multicolumn{5}{c}{\textbf{Silhoutte score}}\\
 &&&& \multicolumn{5}{c}{\textbf{clusters}} &&&
  \multicolumn{5}{c}{ \textbf{clusters}}\\
  
  && \textbf{ Dataset}  && $ 2 $  && $3 $  && $4 $  &&& $2$ && $3$ && $4$  \\
    \bottomrule\toprule
Method && DAE  && $0.5673$  && $0.3195$  && $0.3195$  &&&$0.5312$ && $0.5249$ &&$0.4958$    \\
&&  Miss Forest  && $0.6184$  && $0.5106$  && $0.5106$   &&&$0.5295$ && $0.5212$&&$0.5058$    \\
&&  MICE  && $0.6166$  && $0.5135$  && $0.5135$    & && $0.5296$ && $0.5210$&&$0.5068$\\
&& KNN  && $0.5721$  && $0.3071$  && $0.3071$ &&&  $0.5319$&&$0.5238$&&$0.4921$ \\
\bottomrule
    \end{tabular}}
    
    \vspace{1cm}
    
     \centering
    \caption{Direct analysis of imputation comparing the DAE imputed datasets across missingness degrees.}
    \label{tab:direct:analysis:dae}
    \begin{tabular}{lrrrr}
    \toprule   
    \textbf{Dataset} &  \textbf{Missing} &    \textbf{RMSE} &     \textbf{R2} &                    \textbf{MAPE} \\
    \bottomrule
    \toprule
            DAE &       10 & 547.158 & -1.547 & $1.075 \times 10^{18}$ \\
            DAE &       20 & 601.237 & -0.756 &  $9.531 \times 10^{17}$ \\            
            DAE &       30 & 239.679 &  0.537 &                   $5.429 \times 10^{0}$ \\
            DAE &       40 & 694.770 & -0.511 &  $7.125 \times 10^{17} $ \\
    \bottomrule
    \end{tabular}
    
\end{table}

Consequently, we decided to investigate the performance of the DAE imputation a bit more closely. Therefore, in Table \ref{tab:direct:analysis:dae} we focus on the DAE imputed datasets across all missingness degrees. We make a direct comparison between the actual synthetic values versus the values predicted by DAE by showing values for RMSE, R2 and MAPE as defined in sub-section \ref{subsec:metrics}. The results in this case were not very conclusive in showing any clear pattern on the performance of the method across missingness degrees and is worth investigating a bit further.


\section{Conclusion} \label{sec:conclusion}

Our results on synthetic dataset generation showed that we managed to create a realistic dataset that is similar and comparable to the original dataset based on KMeans analysis of the clustering metrics such as Silhouette score for the two datasets. This was an important step for us because we needed a way to use this dataset with its ground truth values to assess imputation  accuracy, because in real observed datasets, when values are really missing there is no ground truth to determine conclusively how effective a data recovery mechanism is. However, if missingness is simulated, ground truth values exist for all missing values and direct assessment of imputation becomes possible.  Moreover, through classification experiments we were able to have an objective way to compare performance of imputation  for various methods, this was done to allow us the ability to make a recommendation on which method works best for our dataset. We observed that the DAE imputed datasets yielded classifiers that were average performers during training and validation at $87 \%$ and $ 74 \% $ accuracy with minimum log loss of $0.073$, however, when these models were tested on the original observed cleaned sub dataset the models performed way better than the other methods on lower missingness levels of up to $20 \%$ when one considers the loss metric, while on accuracy the performance was slightly lower in comparison to the other methods. On the other hand, some of the state of the art imputation methods such as MICE yielded high accuracy on both the training and testing datasets, however when tested on the unseen clean original sub-dataset the performance was lower. 

Therefore since our interest lies not in general classification accuracy but on how a method performed on a yet unseen clean observed sub-dataset the winning method is the DAE imputer on the loss metric especially on misssingness degrees that are closer to the actual missingness in the observed dataset of about $24 \%$. On the other hand the MICE method is more accurate on all the missingness degrees, even though it is not too significantly different from the other state of the art methods i.e. Miss Forest the Random Forest based imputer method and the K Nearest Neighbour based imputer.

Finally, these results also indicate a possibility that combining methods and improving on model design especially for DAE method may lead to improved overall performance on the original unseen dataset. In conclusion, through, our experiments on synthetic dataset generation, simulation of missing data based on actual data missingness and usage of various missing data imputation methods, we can propose a methodological approach that has the potential to deal with missingness on unseen real datasets. This method allows one to explore various imputation techniques on a realistic synthetic dataset providing an opportunity to assess and determine what method can be most suitable for a given missingness degree or missingness scheme in some observed dataset. 

\bibliographystyle{splncs04nat}
\bibliography{library}

\end{document}